\PassOptionsToPackage{table,xcdraw,dvipsnames}{xcolor}
\documentclass[10pt,twocolumn,letterpaper]{article}

\usepackage{iccv}              

\definecolor{iccvblue}{rgb}{0.21,0.49,0.74}
\usepackage[pagebackref,breaklinks,colorlinks,citecolor=iccvblue, urlcolor=red]{hyperref}
\definecolor{maroon}{cmyk}{1,0,1,0.5}
\def\paperID{15533} 
\def\confName{ICCV}
\def\confYear{2025}

\title{Preserve Anything: Controllable Image Synthesis with Object Preservation}

\author{Prasen Kumar Sharma \quad Neeraj Matiyali \quad Siddharth Srivastava \quad Gaurav Sharma \vspace{0.1em}\\
{\normalsize Typeface}
\vspace{0.1em}\\
{\texttt{\normalsize\{prasen.sharma, neeraj.matiyali, siddharth, gaurav\}@typeface.ai}}
}

\usepackage[accsupp]{axessibility}  

\begin{document}
\maketitle
\begin{abstract}
We introduce \textit{Preserve Anything}, a novel method for controlled image synthesis that addresses key limitations in object preservation and semantic consistency in text-to-image (T2I) generation. Existing approaches often fail (i) to preserve multiple objects with fidelity, (ii) maintain semantic alignment with prompts, or (iii) provide explicit control over scene composition. To overcome these challenges, the proposed method employs an N-channel ControlNet that integrates (i) object preservation with size and placement agnosticism, color and detail retention, and artifact elimination, (ii) high-resolution, semantically consistent backgrounds with accurate shadows, lighting, and prompt adherence, and (iii) explicit user control over background layouts and lighting conditions. Key components of our framework include object preservation and background guidance modules, enforcing lighting consistency and a high-frequency overlay module to retain fine details while mitigating unwanted artifacts. We introduce a benchmark dataset consisting of 240K natural images filtered for aesthetic quality and 18K 3D-rendered synthetic images with metadata such as lighting, camera angles, and object relationships. This dataset addresses the deficiencies of existing benchmarks and allows a complete evaluation. Empirical results demonstrate that our method achieves state-of-the-art performance, significantly improving feature-space fidelity (FID 15.26) and semantic alignment (CLIP-S 32.85) while maintaining competitive aesthetic quality. We also conducted a user study to demonstrate the efficacy of the proposed work on unseen benchmark and observed a remarkable improvement of $\sim25\%$, $\sim19\%$, $\sim13\%$, and $\sim14\%$ in terms of prompt alignment, photorealism, the presence of AI artifacts, and natural aesthetics over existing works.  
\end{abstract}    
\section{Introduction}
\label{sec:intro}

Controlled Image Synthesis (CIS), with an emphasis on preserving target objects in text-to-image (T2I) generated images, has emerged as an important area of research with widespread applications. A prominent use case lies in e-commerce marketing, where products are showcased in diverse environments with varying backgrounds, lighting, and other conditions to enhance consumer engagement. Early work in CIS focused mainly on the use of linguistic cues using frameworks such as Stable Diffusion (SD) \cite{Rombach_2022_CVPR}, while more recent methods, including ControlNet \cite{Zhang_2023_ICCV}, have incorporated additional image-based guidance to improve control over the generation process \cite{Zhang_2023_ICCV, controlnet_plus_plus, qin2023unicontrol, sun2024anycontrol}.

Despite these advances, existing approaches face several significant challenges that limit their effectiveness. Methods relying on linguistic inputs or simple overlays often fail to preserve the integrity of target objects when blended with generated scenes. For example, ControlNet \cite{Zhang_2023_ICCV} requires overlaying target objects onto generated backgrounds, followed by automated refinement (\eg, using SDXL-Refiner \cite{podell2024sdxl}) or manual editing to achieve visual realism. Such post-processing steps introduce additional complexity and dependencies on external tools, \eg, Adobe Photoshop. Similarly, methods that rely on pre-trained, customized models \cite{yu2023inpaint, insertdiffusion} often lack flexibility, particularly for generating high-resolution images, and may produce suboptimal results due to the limitations of these models.

Another key limitation of existing methods lies in their inability to achieve seamless visual harmony between the foreground and background. Techniques such as Replace Anything \cite{chen2024virtualmodel} fail to consider the visual appearance of target objects in the context of the generated scene, leading to unrealistic compositions. Advanced approaches like AnyScene \cite{anyscene}, while capable of producing impressive results, rely on Poisson blending \cite{poissonblending}, which is inadequate for preserving high-frequency details such as text or fine patterns on objects. Subject-driven methods, including DreamBooth \cite{ruiz2023dreambooth} and Custom Diffusion \cite{kumari2022customdiffusion}, require extensive fine-tuning on datasets of similar objects, which is time-consuming and prone to introducing artifacts, particularly when multiple objects are composited within the same scene. Furthermore, these methods provide limited control over the structural layout of the background and lack explicit mechanisms for incorporating geometric guidance such as lighting direction. These constraints hinder their ability to generate scene-specific images tailored to specific applications. Additionally, the datasets used to train CIS models often lack comprehensive annotations, including lighting directions, camera parameters, and aesthetic attributes, which further limits their utility in guiding image synthesis toward high-quality and stylistically consistent outputs.

To address these challenges, we propose a novel framework for CIS, called \textrm{Preserve Anything}, which significantly enhances the ability to preserve target objects while improving the quality and controllability of the generated images. Our method introduces an N-channel extension to ControlNet \cite{Zhang_2023_ICCV}, enabling not only robust object preservation but also the specification of structural layouts for the background and consistent lighting. To further enhance visual realism, we replace traditional Poisson blending \cite{poissonblending} with a lightweight yet effective post-processing method that overlays high-frequency details, such as text and patterns, onto target objects without requiring additional pre-trained models or computationally expensive modules. 

To overcome the limitations of existing datasets, we introduce {PA-240K}, a large-scale benchmark dataset comprising both natural and synthetically rendered images. This dataset is meticulously curated to include diverse prompts, spatial relationships, and geometric cues, such as lighting directions in spherical coordinates and camera parameters, alongside aesthetic scoring. While the real-image subset aids in object preservation, the 3D rendered subset enable precise control over lighting, thereby enhancing the generalizability of our approach. Furthermore, {PA-240K} has the potential to support a wide range of high-level computer vision tasks, such as shadow estimation \cite{shadow_generation}, depth map generation \cite{depthanything}, and surface normal prediction \cite{bae2024dsine}. Through extensive evaluations on traditional benchmarks and user studies conducted on real-world images, we demonstrate the efficacy of our method across a variety of factors, including natural aesthetics, object preservation, and applicability in marketing scenarios. Our results show significant improvements over existing methods, establishing {Preserve Anything} as a robust and versatile framework for CIS. We achieve a significant improvement of $\sim 6\%$ in terms of FID \cite{fid} score cf.\ AnyScene \cite{anyscene}. We also provide a detailed user-study and observe a remarkable improvement of $\sim 25\%$, $\sim 19\%$, $\sim 13\%$, and $\sim 14\%$ in terms of prompt alignment, photorealism, the presence of AI artifacts, and natural aesthetics. The proposed method can be used to seamlessly composite objects into a generated image and allows users to adjust the background layout while ensuring consistent lighting. 
\section{Related Works}
\label{sec:related_works}

\noindent \textbf{T2I Synthesis.} T2I synthesis aims to generate visually coherent images from textual descriptions. Early research predominantly utilized generative adversarial networks (GANs) \cite{pmlr-v48-reed16, stack-gan}, which demonstrated promising results but suffered from challenges such as training instability and limited scalability due to the lack of large-scale datasets. Subsequently, auto-regressive models addressed some of these issues by significantly improving generation quality and robustness. More recently, diffusion-based models \cite{Rombach_2022_CVPR} have emerged as the dominant paradigm, offering state-of-the-art performance. Models such as Stable Diffusion (SD) \cite{Rombach_2022_CVPR}, SDXL \cite{podell2024sdxl}, and other frameworks including FLUX\footnote{\url{https://github.com/black-forest-labs/flux}}, Imagen3 \cite{imagen, imagenteamgoogle2024imagen3}, and DALL-E \cite{pmlr-v139-ramesh21a} have achieved remarkable success in generating high-quality images. In this work, we limit our scope to SD \cite{Rombach_2022_CVPR} and SDXL \cite{podell2024sdxl}, leveraging their strengths in scalable and flexible T2I synthesis.

\noindent \textbf{Controlled Image Synthesis.}  
Controlled image synthesis (CIS) extends T2I capabilities by introducing additional control mechanisms to guide image generation. Techniques such as ControlNet \cite{Zhang_2023_ICCV}, T2I-Adapter \cite{t2i_adapter}, IP-Adapter \cite{ye2023ipadaptertextcompatibleimage}, UniControlNet \cite{unicontrolnet}, and Gligen \cite{li2023gligen} incorporate auxiliary inputs, such as segmentation masks, edge maps, or depth information, to enable fine-grained control over the generated content. More recent advancements, including CocktailNet \cite{hu2023cocktail}, ControlNet++ \cite{controlnet_plus_plus}, and AnyControl \cite{sun2024anycontrol}, further refine these approaches by integrating multimodal inputs, enabling precise control over object placement, style, and layout. These methods have significantly advanced the applicability of CIS across various domains, including product design, advertising, and visual content generation.

\noindent \textbf{Image Inpainting/Outpainting.}  Image inpainting \cite{inpainting} recovers the masked regions of an image with the semantic information of unmasked regions, which can sometimes distort foreground elements or unintentionally regenerate features from the foreground description. Whereas image outpainting \cite{outpainting} extends the image with the background texture
and partially observed objects towards specific directions. In contrast, our approach uses a foreground mask as input, prompting the model to generate a scene that seamlessly blends the foreground with the background. Therefore, this task doesn’t fit neatly into either the inpainting or outpainting categories.
\begin{figure*}[t]
    \centering
    \includegraphics[clip, trim=0cm 0cm 0cm 0cm, width=0.85\linewidth]{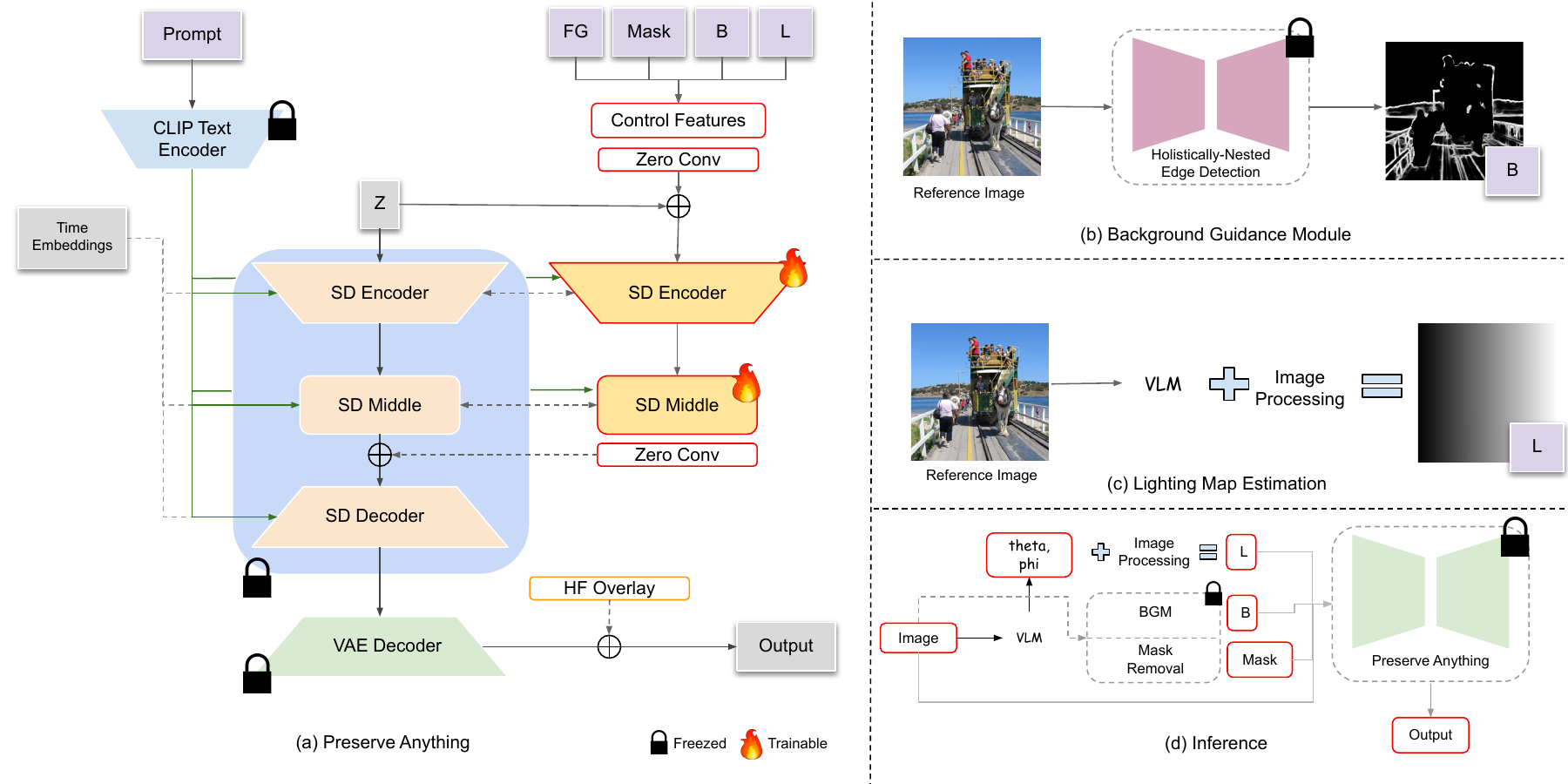}
    \vspace*{-6mm}
    \caption{{\textbf{Overview of the proposed framework}}. (a) \textrm{Preserve Anything} builds upon the {ControlNet} framework \cite{Zhang_2023_ICCV}, extending it to an {N-channel} design with {{N=6}} in the current implementation. The first three channels are dedicated to the {RGB image of the target object}, which is to be preserved against a white background, referred to as {FG}. The fourth channel, labeled Mask, encodes the {binary mask of the object}. The fifth channel represents the {background layout, denoted as B}, which can either be specified by the user or predicted via the {Background Guidance Module (BGM)}, as illustrated in (b). The BGM sub-module processes a reference image to generate a {holistically-nested edge (HED) map}, which serves as the background layout (B). The final channel is reserved for the {lighting cue (L)}, which can either be provided by the user in spherical coordinates or predicted using a vision-language model, as shown in (c). The inference stage is shown in (d), where {Mask is predicted using BiRefNet \cite{zheng2024birefnet} (Mask Removal)} if not provided.}
    \label{fig:main}
\end{figure*}

\noindent \textbf{Object Preservation.}  
Object preservation in CIS focuses on ensuring the integrity and visual realism of target objects within synthesized images. InpaintAnything \cite{yu2023inpaint} introduces a two-step approach that first removes existing objects in the region of interest and subsequently integrates the target object without altering the surrounding scene. ReplaceAnything \cite{chen2024virtualmodel}, tailored for marketing applications, generates high-quality images depicting specified products showcased by humans. This method utilizes user-provided inputs, including the product image, human pose, and textual description. InsertDiffusion \cite{insertdiffusion} employs models like SD 2.1 and SDXL-Refiner \cite{podell2024sdxl} to composite target objects onto existing backgrounds, achieving visually consistent results. AnyScene \cite{anyscene} leverages ControlNet \cite{Zhang_2023_ICCV} by utilizing a target object on a white canvas along with its mask to control object layout in the generated scene. Low-rank adaptation (LoRA)-based approaches \cite{ruiz2023dreambooth, kumari2022customdiffusion, li2023blipdiffusion} utilize few-shot fine-tuning to preserve the characteristics of target objects during generation. These methods, while effective, often require significant computational resources and careful fine-tuning to ensure the fidelity of the synthesized images.

Despite the progress made by these approaches, challenges remain in ensuring seamless integration of target objects with the background, maintaining high-frequency details, and achieving controllable layouts under varying environmental conditions. Our work builds upon these advancements, addressing these limitations to achieve more robust object preservation and enhanced control in image synthesis.
\section{Proposed Method}
\label{sec:propose_method}
We propose a novel framework, {Preserve Anything}, which synthesizes high-quality images with explicit object preservation and enhanced user control over scene composition. The method integrates a carefully curated dataset, an extended ControlNet architecture, and specialized modules for object preservation, background guidance, and high-frequency detail reintegration. Below, we describe the dataset preparation, framework components, and training strategy in detail.

\subsection{PA-240K: Data Curation Pipeline}
\label{sec:dataset}
High-quality datasets are crucial for training models capable of robust object preservation and realistic scene synthesis. Existing approaches, such as AnyScene \cite{anyscene}, rely on randomly sampled datasets and captions generated by BLIP2 \cite{blip2}, which often lack adequate aesthetic diversity and contextual richness. To address these limitations, we curate a dataset with two subsets: a Real-Image subset for training object preservation and a 3D-Rendered subset for enforcing consistent lighting and shadow cues.

\noindent \textbf{Real-Image Subset.}  
Images are sourced from MS-COCO \cite{mscoco}, OpenImagesV7 \cite{openimagesv7}, and FFHQ \cite{ffhq}, but unlike prior work, we prioritize aesthetic quality during selection. First, we remove colorless images by evaluating absolute differences between color channels in \textit{lab} color space. Second, we filter images using LAION-Aesthetic scores \cite{schuhmann2022laionb}, retaining only those above a threshold of 5.0, chosen based on the high-quality standards of MIT Adobe 5K \cite{fivek}. 

For each selected image, we use VLM (GPT4o \cite{OpenAI_GPT4_2023}) to generate detailed annotations, including:
(i) five captions of varying lengths (20-50 words) describing objects, background elements, and environmental details such as lighting,
(ii) lighting directions in spherical coordinates (elevation $\theta$ and azimuth $\phi$),
(iii) camera angles and orientations,
(iv) object descriptions with spatial relationships and interactions,
(v) aesthetic scores based on composition, memorability, and marketing potential.

\noindent \textbf{3D-Rendered Subset.}  
This subset consists of 18K rendered images created using Blender’s Cycles renderer \cite{blender}. Scenes are generated using 14 assets and 48 environments from Poly-Haven, with each pose rendered under four lighting configurations. For each render, we save maps capturing the essential 3D geometry, material, and illumination cues, which include albedo, diffuse shading, reflections, shadows, surface normals, and depth. These maps allow our method to learn lighting and shadow consistency and provide a basis for improving the realism of the generated images. The 3D-rendered subset is used for pre-training to capture lighting and shadow cues, while the real image subset is used for fine-tuning object preservation and scene synthesis.

\begin{figure*}[t]
    \centering
    \includegraphics[clip, trim=0cm 3cm 7cm 0cm, width=0.85\linewidth]{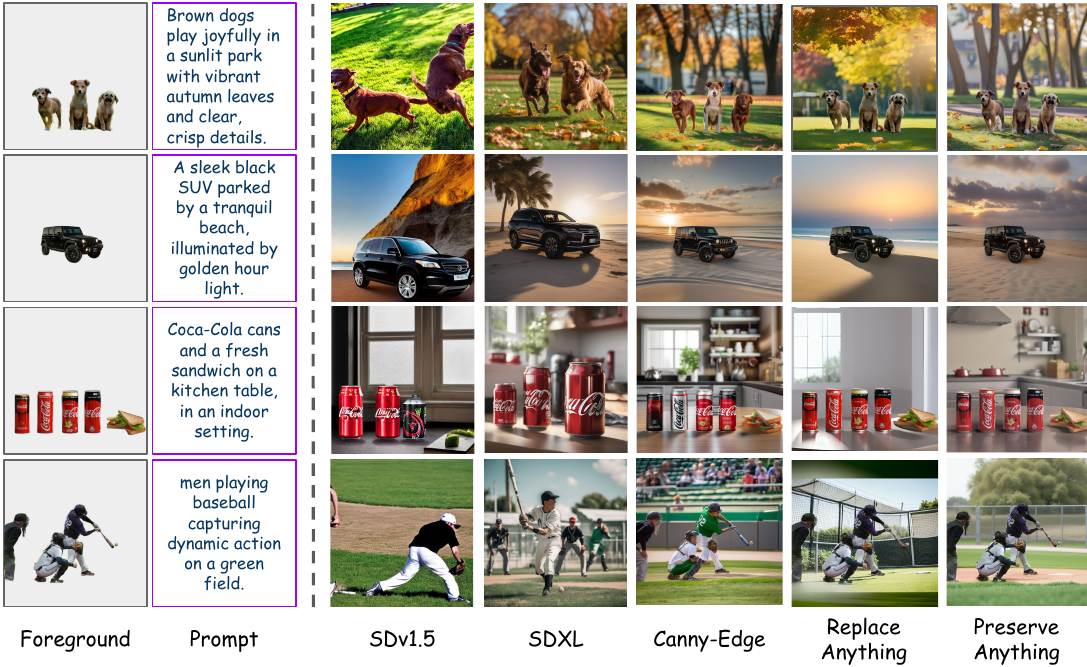}
    \vspace*{-4mm}
    \caption{Qualitative comparison of the proposed method (\textrm{Preserve Anything}) against the SD/ControlNet baselines and the state-of-the-art method $\textrm{Replace Anything}$.}
    \label{fig:main-results}
\end{figure*}

\subsection{Framework Overview}
\label{sec:framework}
Our framework extends ControlNet \cite{Zhang_2023_ICCV} to an N-channel design, enabling the integration of multiple input conditions, such as object masks, background layouts, and lighting maps. It consists of three key modules: the Object Preservation Module, Background Guidance Module, and High-Frequency Overlay Module, shown in Figure~\ref{fig:main}.

\noindent \textbf{Object Preservation Module.}  
This module, denoted $\mathcal{O}$, is responsible for preserving the visual appearance and structural integrity of target objects. The default ControlNet guidance maps (\eg, Canny edges) are insufficient for preserving object appearance because they focus only on structural outlines. To overcome this, we extend ControlNet with an N-channel design, where each condition (\eg, object mask, lighting) is provided as a dedicated input channel.

Input conditions include the target object on a white canvas ($I \in \mathbb{R}^{h \times w \times 3}$), its binary mask ($M \in \mathbb{R}^{h \times w \times 1}$), optional background layout ($B \in \mathbb{R}^{h \times w \times 1}$), and optional lighting map ($L \in \mathbb{R}^{h \times w \times 1}$). These inputs are encoded into a conditioning vector $c^f$ via a lightweight CNN encoder $\mathcal{E}$:
\begin{equation}
c^f = \mathcal{E}(I, M, B, L).
\end{equation}
The conditioning vector $c^f$ is fused into the latent space of the diffusion model to guide generation. The use of object mask $M$ ensures color and detail fidelity for the target object without requiring additional loss functions.

\noindent \textbf{Background Guidance Module.}  
This module allows users to specify the layout of the background scene. During training, background layouts ($B$) are derived from reference images ($R$) using Holistic-Nested Edge Detection (HED) \cite{hed}, capturing the structural appearance of the background:
\begin{equation}
B = \mathcal{H}(R).
\end{equation}
Foreground regions in $B$ are masked out using $M$, ensuring clear separation between objects and the background. If no layout is provided, $B$ defaults to an empty channel filled with zeros. This ensures the framework remains robust in scenarios where explicit background control is not needed.

\noindent \textbf{Lighting Maps.}  
To enforce consistent lighting, we introduce an additional channel $L$ representing lighting direction. Lighting directions are represented by azimuthal angles ($\phi$) and converted into gradient maps simulating illumination intensity. These maps align shadows and highlights with real-world conditions. 

\noindent \textbf{High-Frequency Overlay Module.}  
This module reintegrates high-frequency details, such as text and patterns, onto the target object in the synthesized scene. Given the generated scene $J$ and the target object image $I$, we combine the low-frequency details $\ell_{lf}$ from $J$ with high-frequency details $\ell_{hf}$ from $I$:
\begin{equation}
{\hat{J}} = \ell_{lf}(J) + \ell_{hf}(I).
\end{equation}
This heuristic approach avoids reliance on external blending tools and ensures seamless reintegration of fine details. {More details are available in the supplementary material.}

\subsection{Training Strategy}
\label{sec:training}
The training process involves two stages. First, the model is pre-trained on the 3D-rendered subset to capture lighting and shadow cues. Second, it is fine-tuned on the curated Real-Image subset for object preservation and scene synthesis. To enhance robustness, we randomly prune input channels ($B$, $L$) during training, enabling the model to generalize to scenarios where these cues are absent. We also use classifier-free guidance by randomly dropping textual prompts and conditions at a rate of 0.05. The framework is initialized with a pre-trained Canny-ControlNet checkpoint \cite{Zhang_2023_ICCV} for efficiency. We also extend experiments to SDXL \cite{podell2024sdxl} for generating high-resolution outputs. 

\section{Experimental Setup}
\label{sec:experiments}
We conducted all training experiments on a single NVIDIA A100 GPU with a batch size of 16, using resolutions of $512 \times 512$ and $1024 \times 1024$. The models were trained using the AdamW optimizer \cite{loshchilov2018decoupled} with a learning rate of $1\times10^{-4}$. Training for Stable Diffusion v1.5 (SD v1.5) and SDXL was completed in approximately 40 hours, corresponding to 100K and 10K training steps, respectively. Our training set comprise of 240K images, whereas we perform evaluation on 3K images. Following \cite{anyscene}, these images are a subset of MS-COCO \cite{mscoco}.

To evaluate the performance of the proposed framework against existing methods, we utilized a comprehensive suite of image quality metrics. Visual quality was assessed using Fréchet Inception Distance (FID) \cite{fid}. To measure text-visual consistency, we employed CLIP-Scores \cite{clip}. Additionally, the aesthetic quality of the generated images was evaluated using Neural Image Assessment (NIMA) \cite{nima}, No-Reference Quality Metric (NRQM) \cite{nrqm}, and LAION Aesthetic scores \cite{schuhmann2022laionb}. More details on experimental settings, dataset, and network architecture are available in the supplementary.

\begin{figure}
    \centering
    \includegraphics[clip, trim=0cm 6cm 15cm 0cm, width=0.8\linewidth]{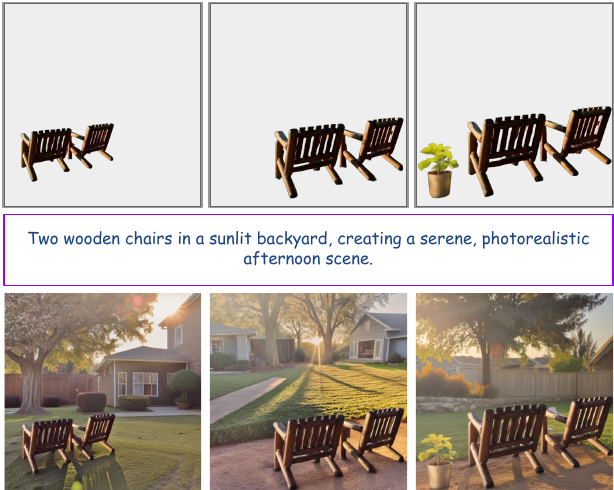}
    \vspace*{-3mm}
    \caption{Our proposed method supports multiple objects and is size and placement agnostic of objects, \ie, given the same prompt, \textrm{Preserve Anything} can generate images with the same objects but having different sizes and placements in the scene.}
    \label{fig:multiple-size-placement}
\end{figure}

\begin{figure}
    \centering
    \includegraphics[clip, trim=0cm 5.5cm 6cm 0cm, width=0.75\linewidth]{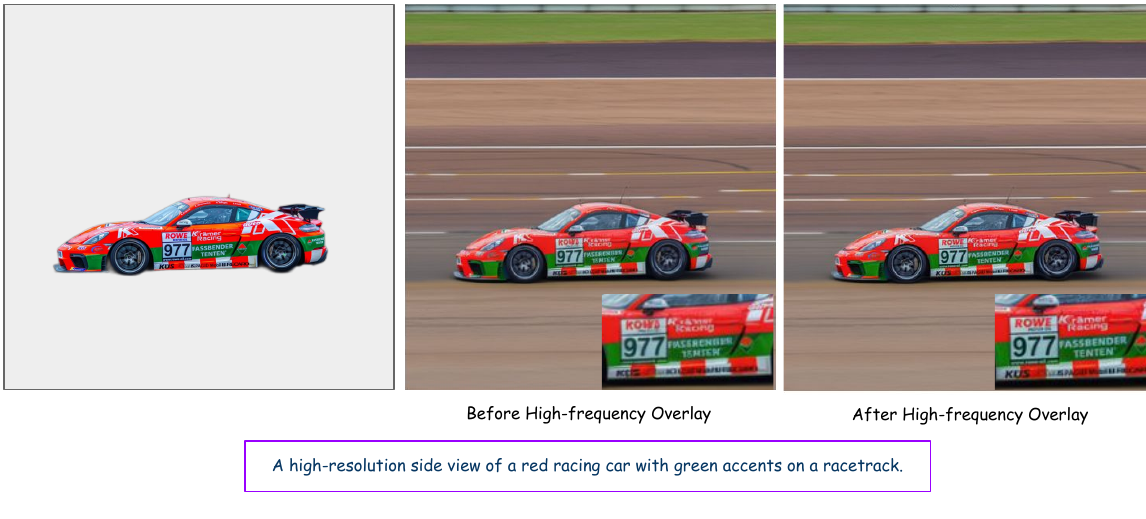}
    \vspace*{-4mm}
    \caption{High-frequency overlay module recovers complex text in the generated images.}
    \label{fig:text-overlay}
\end{figure}

\begin{figure}
    \centering
    \includegraphics[clip, trim=0cm 6cm 15cm 0cm, width=0.75\linewidth]{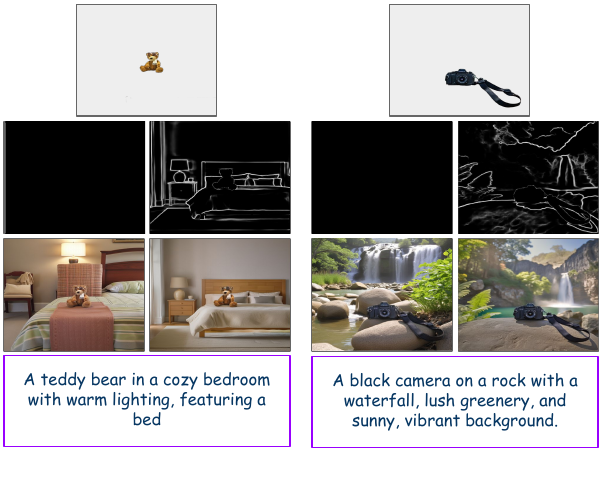}
    \vspace*{-8mm}
    \caption{Our method allows control over the background appearance via an edge map. The left column shows generated image without background control, while the right column shows generated with background control.}
    \label{fig:background}
\end{figure}

\begin{figure}
    \centering
    \includegraphics[clip, trim=1cm 5cm 1cm 0cm, width=0.9\linewidth]{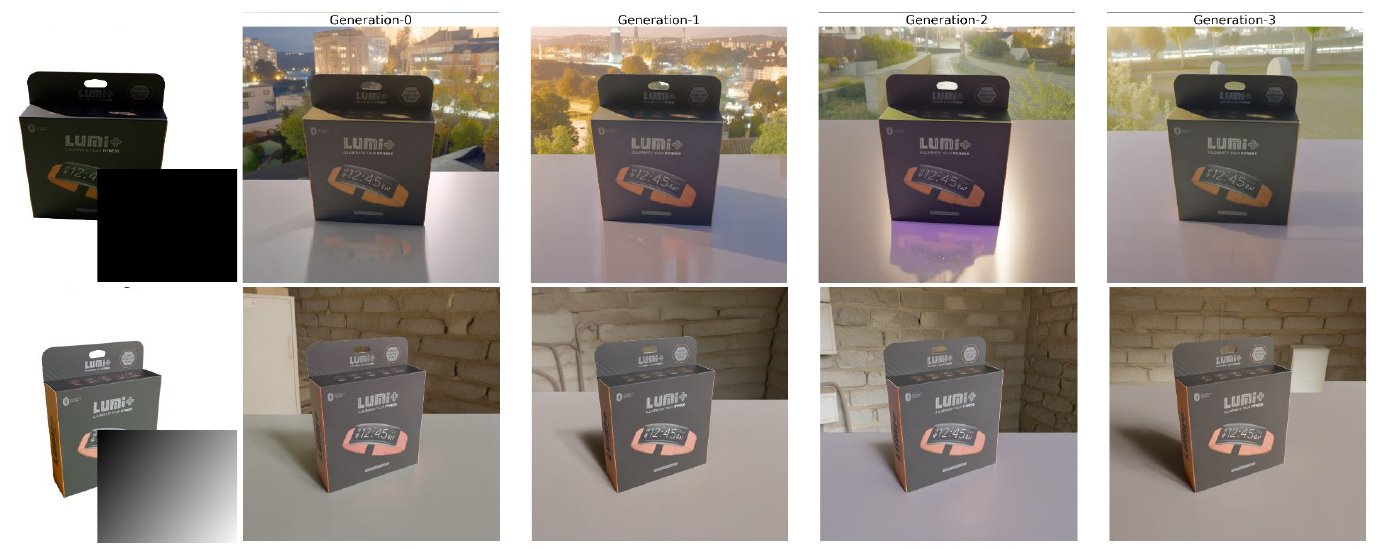}
    \vspace*{-3mm}
    \caption{With explicit lighting direction control via a gradient map (bottom row), our method can generate consistent illumination and shadow in the synthesized image.}
    \label{fig:lighting}
\end{figure}

\begin{table}[t]
    \aboverulesep=0ex
    \belowrulesep=0ex
    \resizebox{\linewidth}{!}{
    \centering
    \begin{tabular}{l|c|cc|ccc}
        \toprule
        & \multicolumn{1}{c|}{\textbf{Feature Space}} & 
        
        \multicolumn{2}{c|}{\textbf{Prompt Alignment}} & 
        
        
        \multicolumn{3}{c}{\textbf{Aesthetic Quality}} \\
        \cmidrule{2-7}
        \textbf{Method} & \textbf{FID} $\downarrow$ & 
        
        
         \textbf{CLIP-S} $\uparrow$ & \textbf{CLIP-IQA} $\uparrow$ & 
        
        
        \textbf{NIMA} $\uparrow$ & \textbf{NRQM} $\uparrow$ & \textbf{LAION-Aes} $\uparrow$ \\
        \midrule
        
        SD v1.5 & 31.00 & 32.54 & 0.6724 & 5.43 & 7.29 & 6.28 \\
        SDXL & 32.84 & 34.18 & 0.6692 & 5.70 & 6.41 &  6.77 \\
        + Canny Edge & 48.59  & 32.54 & 0.6053 & 5.58 & 7.42 & 6.43 \\
        \midrule
        Inpaint Anything & 17.17 &  \textbf{33.30} & 0.6688 & 5.19 & 7.64 & 6.20 \\
        AnyScene & 16.09 & {15.18} & -    & -     & -     & 5.94 \\
        \midrule
        \rowcolor{maroon!20}
        \textbf{Preserve Anything} & \textbf{15.26} & 32.85 & \textbf{0.6869} &  \textbf{5.25} & \textbf{8.23} & 6.06 \\
        \bottomrule
    \end{tabular}
    }
    \vspace*{-3mm}
    \caption{\textbf{Quantitative comparison} of the proposed method against state-of-the-art approaches. Metrics include {Feature Space} (FID), {Prompt Alignment} (CLIP-S), and {Aesthetic Quality} (LAION-Aesthetic). Lower FID indicates better statistical similarity to real images, while higher CLIP-S reflects superior alignment between textual prompts and generated images. {Our method achieves the best FID (15.26), CLIP-IQA (0.6869), NIMA (5.25), and NRQM (8.23), outperforming competing methods such as AnyScene and Inpaint Anything.} Results demonstrate the efficacy of our framework in generating high-quality, semantically consistent images. The SoTA numbers for baseline methods are taken from AnyScene \cite{anyscene}.}

    \label{tab:main_results}
\end{table}

\section{Main Results}
\label{sec:sota_comparison}

To validate the effectiveness of our proposed framework, we compare its performance against several state-of-the-art (SoTA) methods, including Stable Diffusion \cite{Rombach_2022_CVPR}, Canny Edge \cite{Zhang_2023_ICCV}, Inpaint Anything \cite{yu2023inpaint}, and AnyScene \cite{anyscene}. {Note that Replace Anything \cite{chen2024virtualmodel} is currently accessible only through an online demo\footnote{\url{https://huggingface.co/spaces/modelscope/ReplaceAnything}}, limiting us to visual comparisons.} Table \ref{tab:main_results} reports quantitative results across multiple metrics, evaluating \textit{Feature Space} (FID), \textit{Prompt Alignment} (CLIP-S), and \textit{Aesthetic Quality} (LAION-Aesthetic). These metrics comprehensively assess the generative quality, prompt adherence, and visual appeal of synthesized images.

\noindent $\blacksquare$ \textbf{{Feature Space Analysis.}} 
The Fréchet Inception Distance (FID) evaluates the statistical similarity between generated and real images in feature space. Our method achieves the lowest FID of 15.26, significantly outperforming previous methods such as AnyScene (16.09) and Inpaint Anything (17.17). This improvement reflects our framework's ability to produce images that closely resemble real-world distributions, highlighting the effectiveness of our design in preserving high-quality visual features.

\noindent $\blacksquare$ \textbf{{Prompt Alignment.}}  
We use CLIP-S and CLIP-IQA to assess the alignment between textual prompts and generated images. Our method achieves CLIP-S and CLIP-IQA scores of 32.85 and 0.6869, substantially higher than AnyScene (15.18) and other baseline methods (0.6688), respectively. This result demonstrates the superior semantic coherence of our approach, attributed to the integration of prompt-aware guidance during synthesis. The results indicate that our framework effectively translates linguistic input into visually consistent outputs, a key advantage for text-driven applications.

\noindent $\blacksquare$ \textbf{{Aesthetic Quality.}}  
The aesthetic quality of generated images is evaluated using the LAION-Aesthetic score. {Our method achieves a score of 6.06, on par with AnyScene (5.94) and marginally below InpaintAnything (6.20)}. While this metric highlights our framework's ability to produce visually appealing images, future work could explore fine-tuning aesthetic scoring components to match or exceed the performance of SoTA in this domain. We also achieve better results in terms of NIMA and NRQM \cf Inpaint Anything, which conveys the overall photorealism.

\noindent $\blacksquare$ \textbf{{Overall Performance.}}  
The proposed method demonstrates a consistent improvement over competing methods across critical metrics, particularly in FID and CLIP-S, where the advancements are most pronounced. The ability to balance feature-space fidelity and prompt alignment underscores the versatility of our approach, making it suitable for applications requiring precise text-to-image synthesis.

\noindent $\blacksquare$ \textbf{{Qualitative Results.}} Figures \ref{fig:main-results}, \ref{fig:multiple-size-placement}, \ref{fig:text-overlay}, \ref{fig:background}, \ref{fig:lighting}, we show the qualitative results under various settings. We observe that the prompt adherence increases when a background layout is provided (Figure \ref{fig:background}), while if a light map is provided (Figure \ref{fig:lighting}), the method generates background and slightly adapts the object to adhere to the lighting. This is unlike earlier methods, where the lighting control or background control was only prompt-based. Further, we note that qualitatively, our results are better than competing methods (Figure \ref{fig:main-results}). 
Furthermore, our method can handle multiple objects with varying sizes and placement in the scene, as shown in Figure \ref{fig:multiple-size-placement}. Figure \ref{fig:text-overlay} shows the efficacy of such a simple method to retain back the high-frequency details. It shows that this simple technique is as effective as Poisson Blending and does not incur computational cost.

\begin{table}[!t]
    \aboverulesep=0ex
    \belowrulesep=0ex
    \resizebox{\linewidth}{!}{
    \centering
    \begin{tabular}{l|c|cc|ccc}
        \toprule
        & \multicolumn{1}{c|}{\textbf{Feature Space}} & \multicolumn{2}{c|}{\textbf{Prompt Alignment}} & 
        
        
        \multicolumn{3}{c}{\textbf{Aesthetic Quality}} \\
        \cmidrule{2-7}
        \textbf{Config.} & \textbf{FID} $\downarrow$ & 
        \textbf{CLIP-S} $\uparrow$ 
        & \textbf{CLIP-IQA} $\uparrow$ 
        
        
        & \textbf{NIMA} $\uparrow$ & \textbf{NRQM} $\uparrow$ & \textbf{LAION} $\uparrow$ \\
        \midrule
        RGB Only-BLIP2 & 17.31 & 29.68 & 0.6857  
        & 5.16 & 7.93 & 5.86 \\
        + Aesthetic Filtering & 16.63 & 29.91 & 0.6722 
        & 5.18 & 7.82 & 6.00 \\
        \midrule
        RGB Only-GPT4o  & 19.37 & 32.38 & 0.6832 
        & 5.21 & 7.96 & 6.09 \\
        + Mask (A) & 18.31 & 32.45 & \textbf{0.6971} 
        & 5.25 & 7.84 & \textbf{6.11} \\
        \rowcolor{maroon!20}
        + BG  & \textbf{15.26} & \textbf{32.85} & 0.6869 
        & \textbf{5.25} & \textbf{8.23}  & 6.06 \\
        \midrule
        \rowcolor{maroon!30}
        w/ SDXL  & 14.54  & 33.49 & 0.6246 & 
        4.96 & 8.26 & 5.99 \\
        \bottomrule
    \end{tabular}
    }
    \vspace*{-3mm}
    \caption{\textbf{Ablation study} of the proposed framework. The table evaluates the impact of different configurations on performance metrics: {Feature Space} (FID, LPIPS), {Prompt Alignment} (CLIP-S, CLIP-IQA), and {Aesthetic Quality} (NIMA, NRQM, LAION).}
    \label{tab:ablation_study}
\end{table}

\section{Ablation Study}
\label{sec:ablation_study}

To evaluate the contributions of individual components and configurations in our proposed framework, we conduct an extensive ablation study. Table \ref{tab:ablation_study} presents quantitative results across multiple metrics, categorized into \textit{Feature Space} (FID, LPIPS), \textit{Prompt Alignment} (CLIP-S, CLIP-IQA), and \textit{Aesthetic Quality} (NIMA, NRQM, LAION). Each row in the table represents a specific configuration, incrementally adding components to demonstrate their respective contributions.

\noindent $\bigstar$ \textbf{Baseline-RGB Only.}  
The ``RGB Only" configuration serves as the baseline, representing the model's performance without any additional components. This setup achieves an FID of 17.31 and a CLIP-S score of 29.68. While the baseline provides decent performance, the absence of explicit aesthetic filtering or structural guidance limits its ability to align textual prompts with visual outputs effectively.

\noindent $\bigstar$ \textbf{Effect of Aesthetic Filtering.}  
Adding aesthetic filtering (``+ Aesthetic Filtering") refines the training dataset by prioritizing images with high LAION-Aesthetic scores. This inclusion slightly improves the FID (16.63) and maintains comparable CLIP-S, NIMA, and NRQM scores. These results indicate that curating high-quality datasets positively impacts image synthesis, enhancing overall feature space performance.

\vspace{0.5em}
\noindent $\bigstar$ \textbf{Effect of Background Guidance.}  
Introducing background guidance ``+ BG") leverages structural layout hints to guide scene composition. This results in a noticeable improvement in FID (15.26) and a marginal increase in CLIP-IQA (0.6869). However, the LAION Aesthetic score slightly decreases (6.06), suggesting that background guidance enhances spatial consistency at the potential cost of prompt alignment in some cases.

\vspace{0.5em}
\noindent $\bigstar$ \textbf{SDXL Backbone.}  
The incorporation of the SDXL backbone provides a more expressive latent space for image synthesis. The SDXL-based baseline achieves a better FID score of 14.54 \cf SDv1.5-based baseline. Also, the NRQM score has increased, indicating the overall improvement in spatial resolution.
\section{User Study}
\begin{table}[t]
    \centering
    \aboverulesep=0ex
    \belowrulesep=0ex
    \resizebox{\linewidth}{!}{
    \begin{tabular}{l|cccc}
        \toprule
         \textbf{Method} & \textbf{Prompt Alignment}$\uparrow$ & \textbf{Photorealism}$\uparrow$ & \textbf{AI Artifacts}$\downarrow$ & \textbf{Natural Aesthetics}$\uparrow$\\
         \midrule
         Inpaint Anything \cite{yu2023inpaint} & 2.85 & 2.93 & 2.91 & 3.07 \\
         \rowcolor{maroon!30}
         Preserve Anything & \textbf{3.58} & \textbf{3.50} & \textbf{2.52} & \textbf{3.51}\\
         \bottomrule
    \end{tabular}}
    \vspace*{-3mm}
    \caption{\textbf{Quantitative results of the conducted user study.} We evaluate the generative performance of Inpaint Anything and the proposed work across four parameters, comprehensively judging prompt alignment, photorealism, presence of any AI artifacts, and quality of natural aesthetics such as lighting, shadows, and reflection, \etc. Our proposed approach achieves an overall improvement of $\sim 18\%$ \cf Inpaint Anything across the above-listed parameters.}
    \label{tab:user_study}
\end{table}

\begin{figure*}
    \centering
    \resizebox{0.85\linewidth}{!}{
    \begin{tabular}{cc}
         \includegraphics[clip, trim=1cm 23cm 1cm 1cm, width=0.5\textwidth]{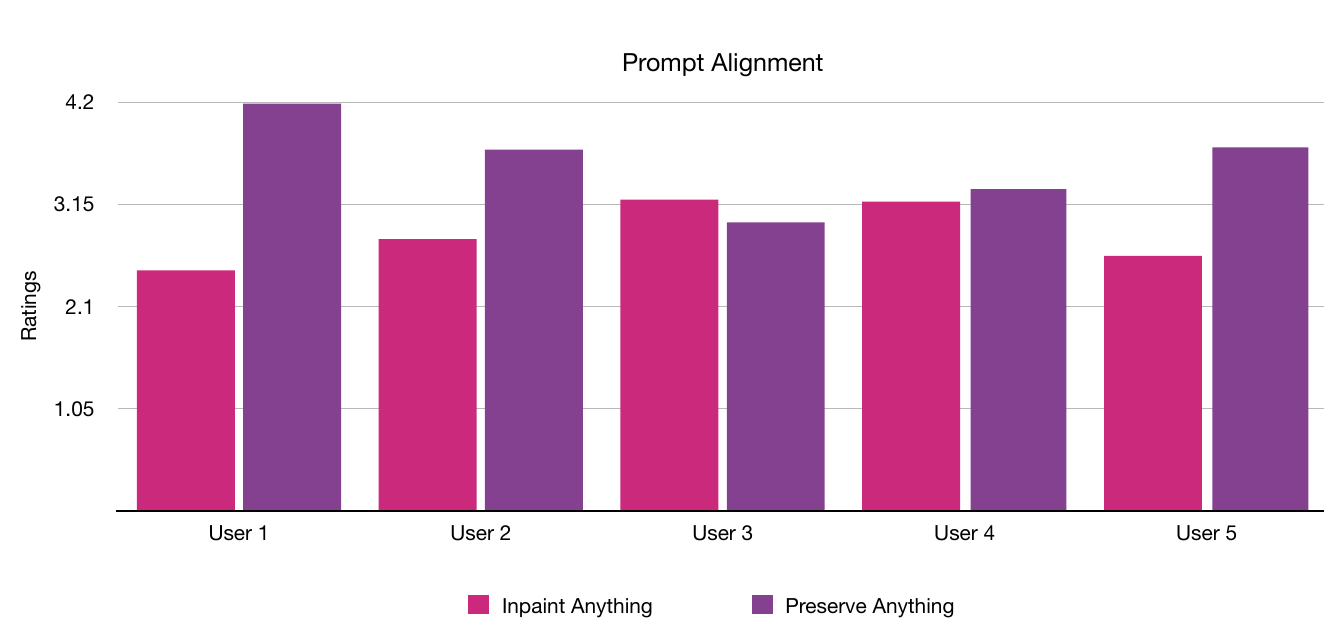} &
         \includegraphics[clip, trim=1cm 23cm 1cm 1cm, width=0.5\textwidth]{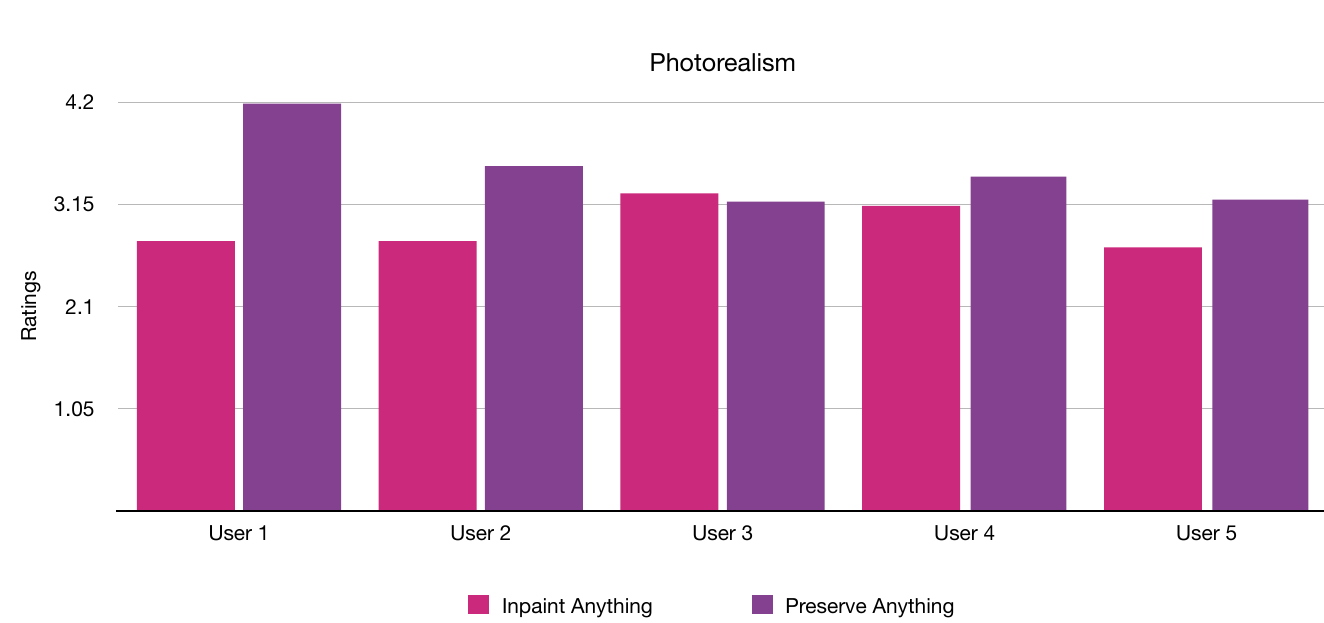} \\

        \includegraphics[clip, trim=1cm 23cm 1cm 1cm, width=0.5\textwidth]{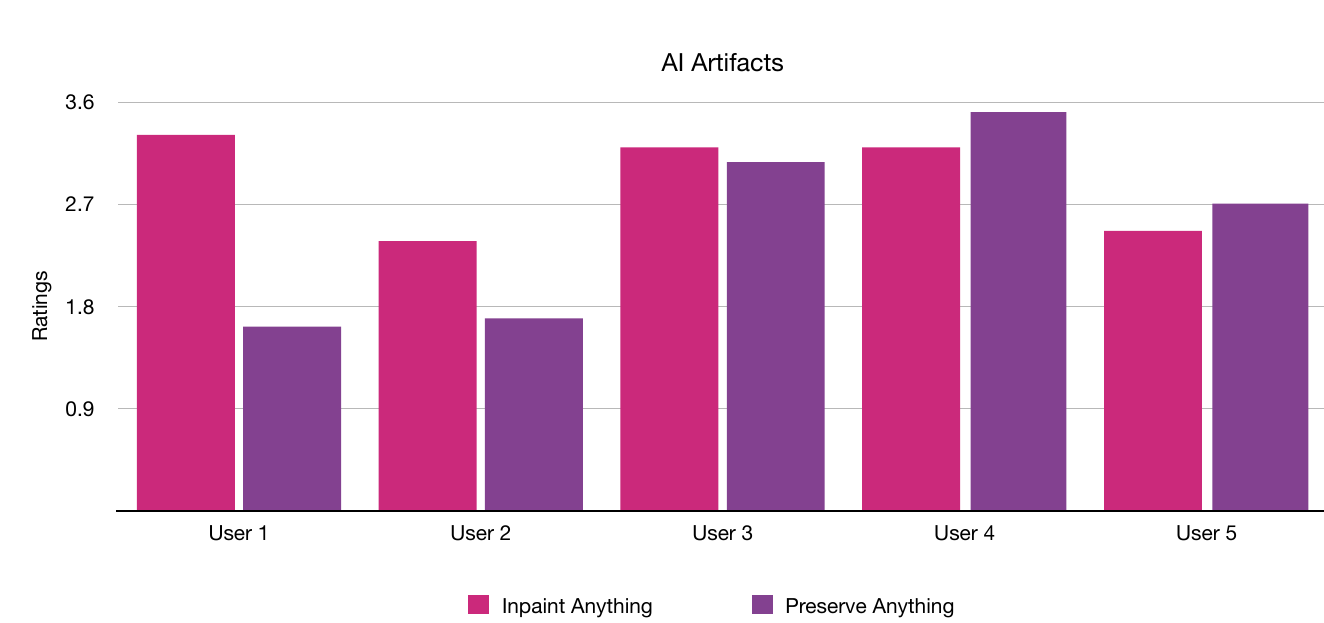} &
         \includegraphics[clip, trim=1cm 23cm 1cm 1cm, width=0.5\textwidth]{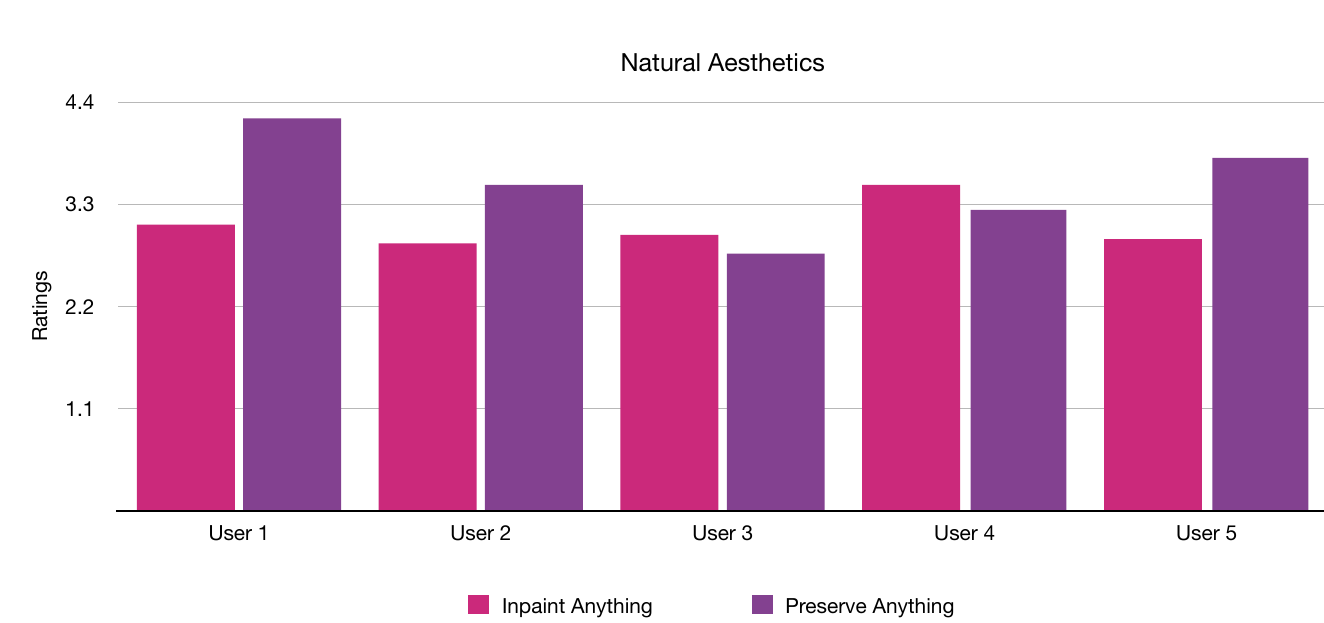} \\
    \end{tabular}}
    \vspace*{-5mm}
    \caption{\textbf{Visual demonstration of user-wise responses across four parameters.} Top-left shows the preferences of users in terms of prompt alignment. Top-right shows the preferences in terms of photorealism. Bottom-left indicates the preferences in terms of AI artifacts. Whereas the Bottom-right depicts the preferences of users related to the quality of natural aesthetics.}
    \label{fig:user_study}
\end{figure*}

\begin{figure*}[ht]
    \centering
    \includegraphics[clip, trim=0cm 4cm 1cm 0cm, width=0.9\linewidth]{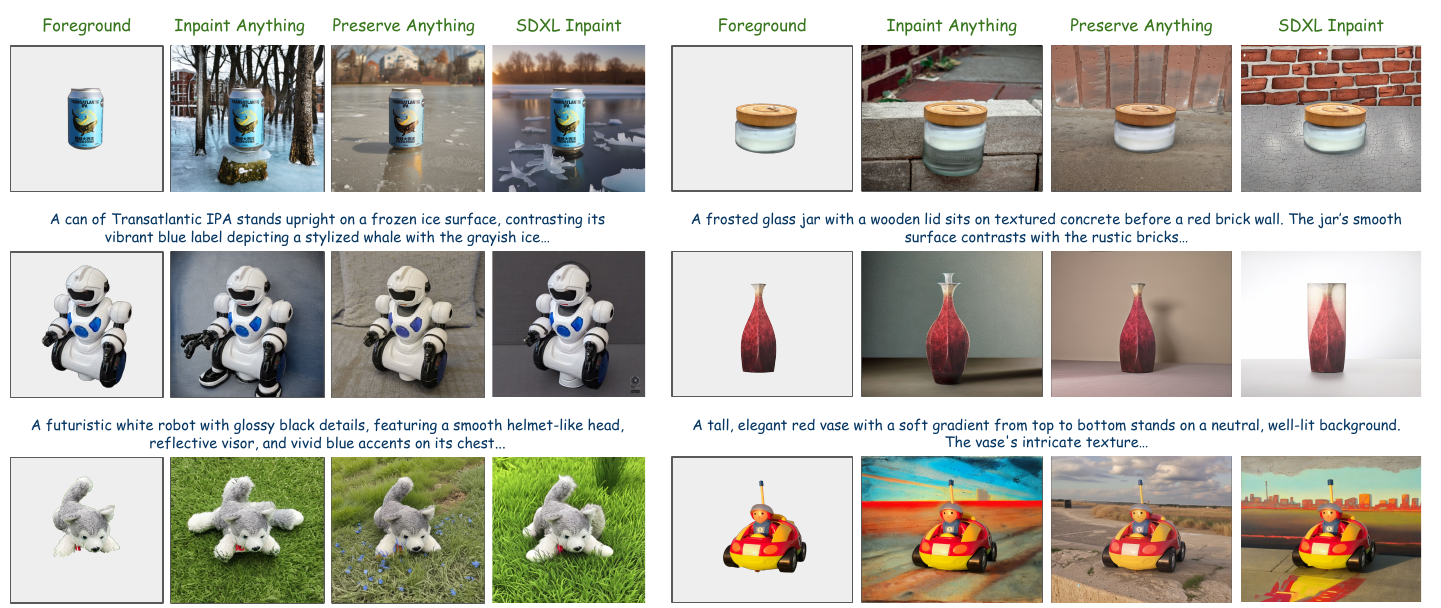}
    \vspace*{-3mm}
    \caption{\textbf{Generalization test.} Qualitative results of Preserve Anything compared to the existing method on the Dreambooth dataset.}
    \label{fig:more_results}
\end{figure*}

To evaluate the generative capability of \textrm{Preserve Anything} and existing methods on CIS, we conducted a user study focused on users' subjective quality assessment. We compare our proposed approach against Inpaint Anything \cite{yu2023inpaint}\footnote{\textcolor{violet}{The code and other implementation details are not publicly available for AnyScene \cite{anyscene}, therefore, we choose Inpaint Anything as our primary baseline for comparison in this user study.}}. We follow a similar process of conducting the user study as described in AnyScene \cite{anyscene}, but with a newer set of questions, targeting important aspects of the perceptual quality assessment of the images. We randomly sample 30 images from Dreambooth \cite{ruiz2023dreambooth} dataset. For each image, we derive the foreground mask of the salient objects to be preserved using BiRefNet \cite{zheng2024birefnet}, and compose them into a composited foreground with a white background. We derive the text prompts by querying the original images to GPT-4o \cite{OpenAI_GPT4_2023}, each length varying from 20 to 50 words. Our curated image-prompt pairs serve as input to each competing method. The generated images using Inpaint Anything and the proposed approach are anonymized and asked to be rated by a group of five annotators from diverse age groups (22 to 50) based on the following criteria:

\begin{itemize}
    \item[$\checkmark$] \textbf{Prompt Alignment:} Rate the semantic alignment between the generated image and text prompt on a scale of 1 to 5. (1-Poor, 5-Best)
    \item[$\checkmark$] \textbf{Photorealism:} Evaluate the image based on its overall realism on a scale of 1 to 5. (1-Poor, 5-Best). We define realism as follows: ``A realistic image resembles a photograph that accurately depicts what you see in real life. It represents objects as they truly are, with precise colors, shapes, and details. When you look at a realistic image, it feels as though you could reach out and touch the objects because they appear so natural and lifelike.'', for user's reference.
    \item[$\checkmark$] \textbf{AI Artifacts:} Rate the presence of nonsensical objects and other AI artifacts in the generated images on a scale of 1 to 5. (1-Best/No artifacts, 5-Poor/Lots of artifacts),
    \item[$\checkmark$] \textbf{Natural Aesthetics:} Evaluate the quality of natural aesthetics such as lighting, shadows, reflection, \etc, \wrt to the generated image on a scale of 1 to 5. (1-Poor, 5-Best).
\end{itemize}

Based on these criteria, the annotators scored the generated images, and the final evaluation results are shown in Table \ref{tab:user_study}. Based on user preferences, it is clear that \textrm{Preserve Anything} surpasses Inpaint Anything in overall generative quality across all four metrics by an average improvement of $\sim 18\%$. In terms of prompt alignment, the proposed method outperforms Inpaint Anything by a significant margin of $\sim 25\%$ in terms of prompt alignment, \ie, images generated using \textrm{Preserve Anything} align more closely with the input prompts compared to Inpaint Anything. In terms of photorealism, we report an average improvement of $\sim 19\%$ over Inpaint Anything in terms of overall photorealism, \ie, the generated images using \textrm{Preserve Anything} exhibit better photorealism \cf Inpaint Anything. Our proposed work shows a significant improvement of $\sim 13\%$ in terms of identifying the presence of AI artifacts in the generated images \cf Inpaint Anything. In terms of natural aesthetics, such as lighting, shadows, reflection, \etc, our proposed work outperforms Inpaint Anything by a significant margin of $~14 \%$, indicating the presence of better natural aesthetics in the generated images. The details of rating per question for each user are shown in Figure \ref{fig:user_study}. 

Overall, we report an average improvement of $\sim 18\%$ over Inpaint Anything in terms of the above-listed parameters. To summarize, the results of the user studies indicate the superior efficacy of \textrm{Preserve Anything} for the task of CIS and its importance in tasks such as poster design and scene personalization, which involve balancing image quality, scene coherence, and object accuracy. The visual comparison on the Dreambooth dataset related to the user study is shown in Figure \ref{fig:more_results}. We observe that the generated images using proposed method are more realistic, with better natural aesthetics like shadow and lighting, \cf existing works.
\section{Conclusions}
\label{sec:conclusion}
We proposed \textrm{Preserve Anything}, a novel framework for controlled image synthesis that addresses limitations in object preservation, prompt alignment, and aesthetic quality. By integrating an N-channel ControlNet, prompt-aware guidance, and modules for object preservation, background guidance, and high-frequency detail reintegration, our approach achieves state-of-the-art performance on various metrics. While our method demonstrates strong semantic alignment and visual fidelity, future work will focus on improving multi-object synthesis, optimizing aesthetic quality, and enhancing computational efficiency for high-resolution applications. Overall, our framework advances T2I generation, offering scalable, user-controllable solutions for high-quality image synthesis.

\newpage
\appendix

\section{Preliminaries}
\label{sec:preliminaries}

Text-to-image (T2I) synthesis methods, such as Stable Diffusion \cite{Rombach_2022_CVPR}, generate images from textual descriptions ($c^t$)  using diffusion models. These models operate in either the pixel or latent domain, with latent-space methods being computationally more efficient. Our work leverages Stable Diffusion, which employs a latent-space formulation for scalable and effective image synthesis. Below, we briefly outline the key components of Stable Diffusion and ControlNet \cite{Zhang_2023_ICCV}, a framework enabling controlled image generation through additional task-specific inputs.

In diffusion models, Gaussian noise $\epsilon \sim \mathcal{N}(0,I)$ is incrementally added to an initial image $x_0$ to produce a noisy sample $x_t$ at timestep $t$, as defined by:
\begin{equation}
x_{t} = \sqrt{\bar{\alpha}_{t}} x_{0} + \sqrt{1-\bar{\alpha}_{t}}\epsilon,
\end{equation}
where $\alpha_t = 1-\beta_t$ and $\bar{\alpha}_t = \prod_{s=1}^{t} \alpha_s$, following a variance schedule $\{\beta_t\}$. A denoising neural network $\epsilon_\theta$ is trained to predict the noise $\epsilon$ by minimizing the objective:
\begin{equation}
\mathcal{L} = \mathbb{E}_{x_{0}, t, \epsilon \sim \mathcal{N}(0,I)} \left[ \left\lVert \epsilon - \epsilon_{\theta}(x_{t}, t, c^t) \right\rVert_2^2 \right].
\end{equation}
Once the denoising network is trained, starting from some random noise $x_T \sim \mathcal{N}(0,I)$, it can be used to sample an image $x_0$ from the learned distribution by iteratively refining  $x_t$:
\begin{equation}
x_{t-1} = \frac{1}{\sqrt{\bar{\alpha}_{t}}} \left( x_t - \frac{\beta_t}{\sqrt{1-\bar{\alpha}_t}} \epsilon_{\theta}(x_{t}, t, c^t) \right) + \sigma_t \epsilon',
\end{equation}
where $\epsilon' \sim \mathcal{N}(0,I)$ and $\sigma_t^2 = \beta_t$. Intermediate estimations of $x_0$ can also be obtained at any timestep $t$ using:
\begin{equation}
\hat{x}_{0} = \frac{x_t - \sqrt{1-\bar{\alpha}_{t}} \epsilon_{\theta}(x_{t}, t, c^t)}{\sqrt{\bar{\alpha}_{t}}}.
\end{equation}
In latent diffusion models (LDMs) (\eg, Stable Diffusion), the forward and reverse diffusion process is done on latent features maps $z = \mathcal E_\text{VAE}(x)$ encoded by a pretrained autoencoder instead of RGB pixels $x$. The denoised latent representation $z_0$ is decoded into the final image using the decoder $x_0 = \mathcal{D}_\text{VAE}(z_0)$.

ControlNet \cite{Zhang_2023_ICCV} extends the controllability of large scale pretrained T2I LDMs process by introducing additional task-specific conditions ($c^f$), such as Canny edges, depth maps, or segmentation maps, which encode structural cues for the target image. The training objective of ControlNet incorporates these additional conditions:
\begin{equation}
\mathcal{L} = \mathbb{E}_{x_{0}, t, c^t, c^f, \epsilon \sim \mathcal{N}(0,I)} \left[ \left\lVert \epsilon - \epsilon_{\theta}(x_{t}, t, c^t, c^f) \right\rVert_2^2 \right].
\end{equation}
This enables fine-grained control over the image generation process, allowing for precise alignment of the target object and structural layout in synthesized images.
\section{Network Details}
{\noindent {\textbf{High-Frequency Overlay.} The goal of this module is to re-establish high-frequency details in the target object of the synthesized image ($J$) from the source image ($I$). To achieve this, we employ a simple mechanism that decomposes an image into its low and high-frequency components. For any image $P$, The low-frequency component ($\ell_{lf}(P)$) is first extracted by applying a Gaussian blur with a kernel of size $17 \times 17$. The high-frequency component is then computed as:

\begin{equation}
\ell_{hf}(P) = P - \ell_{lf}(P).
\end{equation}

We independently compute both the low and high-frequency components for the source and synthesized images, denoted as $\ell_{lf}(I)$, $\ell_{hf}(I)$, $\ell_{lf}(J)$, and $\ell_{hf}(J)$, respectively. Finally, we overlay the high-frequency components of the target object from the source image onto the synthesized image using its binary mask $M$ as follows:

\begin{equation}
\hat{J} = M * \ell_{hf}(I) + (1 - M) * \ell_{hf}(J) + \ell_{lf}(J)
\end{equation}

This approach enables high-frequency detail transfer with minimal computational overhead \cf to complex methods, such as Poisson blending \cite{poissonblending}, while maintaining a favorable trade-off between efficiency and visual quality.}}
\section{Datasets}
\begin{figure*}
    \centering
    \includegraphics[clip, trim=0cm 6cm 0cm 0cm, width=\linewidth]{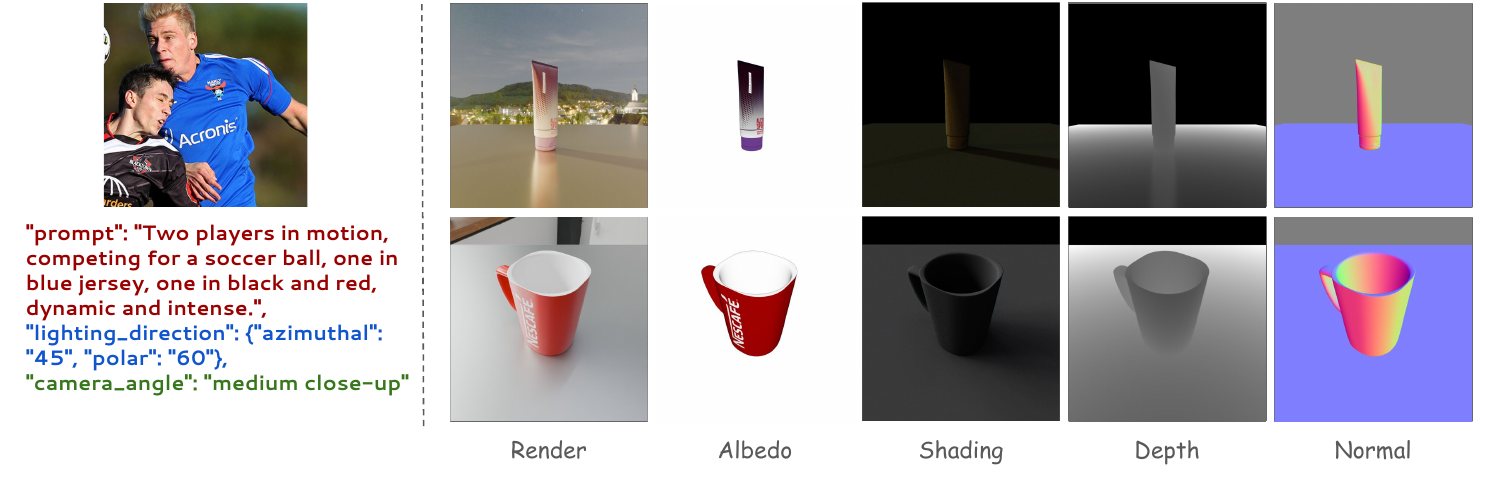}
    \vspace*{-5mm}
    \caption{\textbf{An illustration of our large-scale dataset.} (Left) shows an image from the real-world subset with its derived annotations (prompt and lighting cues for demonstration) from GPT4o. (Right) shows two samples from the curated 3D-rendered synthetic subset with their corresponding albedo, shading, depth, and surface normal maps.}
    \label{fig:dataset_samples}
\end{figure*}

We introduce a novel large-scale dataset for controlled image synthesis (CIS), including real-world and 3D-rendered synthetic images, as outlined in Section 4.1 of the main manuscript. The following subsections provide a detailed description of the curated images and their corresponding annotations, for both real and synthetic image subsets.

\subsection{Real-Image Subset}
Real-world images are sourced from MS-COCO \cite{mscoco}, OpenImagesv7 \cite{openimagesv7}, and FFHQ \cite{ffhq} datasets. However, instead of randomly selecting the images for training CIS models, we propose a filtering strategy to leverage images with better aesthetic quality by using LAION Aesthetic \cite{schuhmann2022laionb} score. Particularly, we utilize images with LAION Aesthetic scores above a threshold of 5.0. The threshold has been set based on the average LAION Aesthetic scores on MIT Adobe 5K \cite{fivek} images. We also discard the grayscale images. 

As reported in Section 4.1 of the main manuscript, for each selected image, we generate detailed annotations using GPT4o \cite{OpenAI_GPT4_2023} as follows:
\begin{itemize}
    \item five captions of varying lengths (20-50 words), describing salient objects and background elements,
    \item dominant light source in terms of spherical coordinates, 
    \item camera orientation, 
    \item objects in the scene with their detailed descriptions, and overall categorical counts, 
    \item spatial relationships between different objects in the scene, \eg, \texttt{The fence is behind the person},
    \item action relationships, \eg, \texttt{The person is holding the coffee cup},
    \item technical details, such as light source with strength and direction, presence of shadows, color palette, depth of field, exposure, camera orientation in terms of roll, pitch, and yaw, and
    \item aesthetic scoring measuring overall composition, focus, clarity, memorability (which indicates ``\texttt{how memorable an image is}''), timelessness, and emotions.
\end{itemize}
Our real-image subset consists of 240K images with the above-detailed annotations. These annotations are beneficial for solving diverse tasks. Quality prompts and lighting conditions are leveraged to enhance photorealism in CIS. Beyond image synthesis tasks, object descriptions with their spatial and action relationships could be used for efficient image understanding, and technical and aesthetic details for image quality assessment. 

\begin{figure*}[t]
    \centering
    \includegraphics[clip, trim=0cm 0cm 5cm 0cm, width=\linewidth]{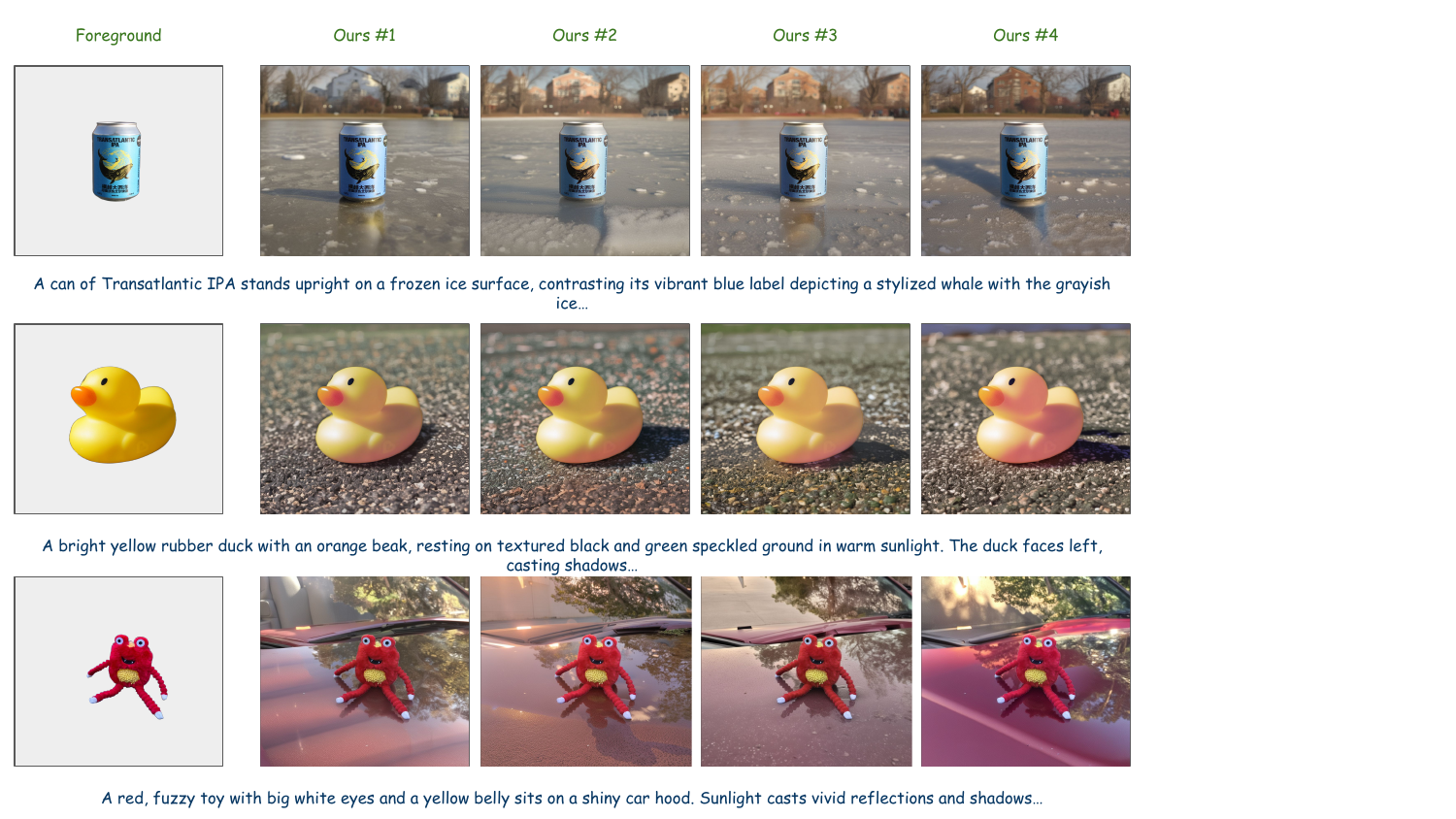}
    \vspace*{-4mm}
    \caption{\textbf{Consistency in natural aesthetics \wrt prompts.} Given a fixed background layout, observe how well the proposed method maintains the natural aesthetics such as lighting, casting shadows, and reflection \wrt input foreground and textual prompt, across various generations.}
    \label{fig:consistency}
\end{figure*}

\subsection{3D Rendered Synthetic Subset}
In addition to the real-image subset, we also construct a synthetic subset to control different natural aesthetics, such as lighting, and shadow, in the generated images. For this, we use Blender's \cite{blender} raytrace-based Cycles renderer to generate images of 3D assets. We collect different HDR environment textures from PolyHaven\footnote{\url{https://polyhaven.com/}} for lighting the 3D assets while rendering. For each asset/environment pair, we vary camera orientation in terms of azimuthal angle from 0$^\circ$ to 360$^\circ$ and elevation angle from 45$^\circ$ to -15$^\circ$.

For each render, we save maps capturing the essential 3D geometry, material, and illumination cues, which include albedo, diffuse shading, reflections, shadows, surface normals, and depth. These maps allow our method to learn lighting and shadow consistency and provide a basis for improving the realism of the generated images. 

Figure \ref{fig:dataset_samples} illustrates some samples from real-image and 3D-rendered subsets. (Left) shows an image from the real-world subset with its derived annotations (prompt and lighting cues for demonstration) from GPT4o. The derived prompt describes the scene well, along with the accurate estimation of lighting coordinates and camera angle. (Right) shows two samples from the curated 3D-rendered synthetic subset with their corresponding albedo, shading, depth, and surface normal maps.
\section{Quality Assessment}
We employ the following image quality assessment metrics to measure the quality of generated images using \textrm{Preserve Anything} against existing works. These metrics evaluate image quality by assessing divergence from real images in the feature domain, prompt adherence, and visual appeal of the generated images. The metrics are briefly described as follows:

\begin{itemize}
    \item The Fréchet Inception Distance (FID)\footnote{\url{https://github.com/GaParmar/clean-fid}} \cite{fid} metric assesses the quality of generated images by quantifying the divergence between feature distributions of real and generated images. It is commonly used to evaluate generative models, especially Generative Adversarial Networks (GANs) \cite{goodfellow2014generative}. It is defined as the distance between two Gaussian distributions-- one representing the real images and the other representing the generated images. These distributions are typically derived from a pre-trained Inception \cite{szegedy2015going} model. A lower FID score indicates greater similarity between generated images and real images, reflecting the better performance of the generative model.

    \item The CLIP\footnote{\url{https://huggingface.co/docs/transformers/en/model_doc/clip}} \cite{clip, clipiqa} scores assess the alignment between an image and a corresponding text prompt by measuring the similarity between their respective feature distributions. A higher CLIP score indicates better semantic correspondence between the image and the textual description.
    
    \item Neural Image Assessment (NIMA) \cite{nima} utilizes a convolutional neural network (CNN) to predict an image's aesthetic score on a scale of 1 to 10, which can be interpreted as a continuous value reflecting the image's quality or classified such as poor, average, excellent, \etc
    
    \item No-reference Quality Metric (NRQM) \cite{nrqm} is designed to predict the perceptual quality of images by analyzing various features, including sharpness, contrast, and other aesthetic or perceptual indicators. Specifically, it derives three types of low-level statistical features in both spatial and frequency domains, to measure the quantum of super-resolved artifacts, and learn a two-stage regression model to predict the quality of images, w/o referring to ground truth images.
    
    \item LAION Aesthetic\footnote{\url{https://github.com/LAION-AI/aesthetic-predictor}} \cite{schuhmann2022laionb} (LAION-Aes) score estimates the aesthetic quality of an image using large-scale pre-trained models. It predicts the aesthetic appeal of an image, aligning with human perceptions of visual attractiveness, on a scale of 1 to 10, with 1 reflecting poor aesthetic quality. It is trained on a comprehensive set of images rated according to human judgment, such as Aesthetic Visual Analysis (AVA) \cite{ava}. The inline model is composed of simple linear layers on top of CLIP ViT/14 \cite{clip}, and typically generates a score that reflects the overall visual appeal of an image, offering insight into how aesthetically pleasing it may be to human viewers.
\end{itemize}

We use pyiqa\footnote{\url{https://github.com/chaofengc/IQA-PyTorch}}, a popular image quality assessment tool, for measuring NIMA and NRQM scores.
\section{Ablation Study}
{We ablate the baseline “RGB Only-BLIP2” from Table 2 in paper by increasing the batch size from 16 to 128. With large batch size, the FID score (16.98) improves slightly (+0.33), while other scores show no significant gains. Larger batches may help stabilize gradients – evident from improved FID , but do not lead to better quality outputs.}

\section{More Results}
The images generated using the proposed method demonstrate enhanced photorealism, preserving natural aesthetics such as lighting and shadows (see Figure \ref{fig:consistency}), while the target objects are well-integrated with the background scene. {Following recent work \cite{aigivc}, and due to lack of established metrics to measure lighting and shadow consistency, we use GPT-4o as a visual critic to rate generated images on shadow strength, direction, and lighting adherence \wrt shadows. The average scores positively favor the pretraining. This shows (which is also evident from Figure \ref{fig:consistency}) that pretraining with 3D rendered subset helps in generating images with consistent lighting and shadows.}

{
    \small
    \bibliographystyle{ieeenat_fullname}
    \bibliography{main}
}

\end{document}